\documentclass[conference]{IEEEtran}
\IEEEoverridecommandlockouts
\usepackage{cite}
\usepackage{amsmath,amssymb,amsfonts}
\usepackage{algorithmic}
\usepackage{url}
\usepackage{soul}
\usepackage{multirow}
\usepackage{color}
\usepackage{graphicx}
\usepackage{textcomp}
\usepackage{xcolor}
\def\BibTeX{{\rm B\kern-.05em{\sc i\kern-.025em b}\kern-.08em
    T\kern-.1667em\lower.7ex\hbox{E}\kern-.125emX}}
\usepackage{caption} 
\begin{document}
%
\title{Multi-views Embedding for Cattle Re-identification}

\author{
\hspace{0.\linewidth}\IEEEauthorblockN{Luca Bergamini\IEEEauthorrefmark{1}}
\IEEEauthorblockA{
\hspace{0.\linewidth}
luca.bergamini24@unimore.it}
\and

\IEEEauthorblockN{Angelo Porrello\IEEEauthorrefmark{1} \hspace{0.15\linewidth}}
\IEEEauthorblockA{
angelo.porrello@unimore.it \hspace{0.15\linewidth}}
\and

\IEEEauthorblockN{ Andrea Capobianco Dondona\IEEEauthorrefmark{2}\IEEEauthorrefmark{3}, Ercole Del Negro\IEEEauthorrefmark{2}\IEEEauthorrefmark{3}, Mauro Mattioli\IEEEauthorrefmark{3}, Nicola D'Alterio\IEEEauthorrefmark{3}, Simone Calderara\IEEEauthorrefmark{1}}
\\

\IEEEauthorblockA{ \IEEEauthorrefmark{2}Farm4Trade Srl, Chieti, Italy \\  \IEEEauthorrefmark{1}AImageLab, University of Modena and Reggio Emilia, Modena, Italy\\
 \IEEEauthorrefmark{3}Istituto Zooprofilattico Sperimentale dell'Abruzzo e del Molise 'G.Caporale', Teramo, Italy}
}

\maketitle
\begin{abstract}
People re-identification task has seen enormous improvements in the latest years, mainly due to the development of better image features extraction from deep Convolutional Neural Networks (CNN) and the availability of large datasets. However, little research has been conducted on animal identification and re-identification, even if this knowledge may be useful in a rich variety of different scenarios. Here, we tackle cattle re-identification exploiting deep CNN and show how this task is poorly related with the human one, presenting unique challenges that makes it far from being solved. We present various baselines, both based on deep architectures or on standard machine learning algorithms, and compared them with our solution. Finally, a rich ablation study has been conducted to further investigate the unique peculiarities of this task.

\end{abstract}


\section{Introduction}

\subsection{Animal Re-identification Motivations}
Animal Re-identification shares some of the aims of the human task, while also including new challenges. The identification process represents a pillar for national and international trade, especially for animals representing crucial economic assets. Furthermore, it constitutes a method for validating the quality and the "authenticity" of the animal being traded. Similarly, for animals supplying products intended for human consumption, the animal identity and the traceability along the entire value chain are prerequisites for the certification of the quality and the safeness of the product for final consumers. In fact, as some of these animals may host and transmit pathogens, a monitoring system is essential to avoid the spread of such diseases to humans and animals and it is necessary to easily identify and track the origin of infected products. Finally, stock theft represents an issue that often outbreaks into a social challenge in developing countries. As an example, India reported almost ten thousand cattle thefts in 2015 while the number of horse theft has grown past forty thousands world wide \cite{cattlewiki}, \cite{cattleindia}.

On the other hand, re-identification systems for pets has seen some interest in the computer vision community \cite{moreira2017my}, mainly aiming to retrieving lost "family members".

Finally, re-identification systems may represent an opportunity for safeguarding endangered species, acting as a crucial aid for studying wildlife and for conservation actions. For such animals, traditional identification systems make use of electronic chips placed in collars and require the animal to be captured and immobilised at least once. Such practice may be unfeasible for aggressive or elusive species, and typically requires GSM or satellite transmitters, the latters being very expensive and often impractical. Again, only few noticeable novel and unobtrusive methods \cite{crouse2017lemurfaceid}, \cite{deb2018face}, \cite{norouzzadeh2018automatically} have been proposed, mainly due to the lack of large datasets publicly available.

\subsection{Cattle Re-identification Motivations}
The number of cattle in Europe in 2016 stood at almost 122 million, of which 23.3 million were dairy cows \cite{cattleeruope}, and the number is growing past 1400 million in the world based on the latest surveys \cite{cattlestat}. Although  efficient identification and re-identification systems already exists, it is mandatory to develop new tools that can support the existing ones not only to ensure milk and meat safeness, but also to avoid kidnappings and counterfeits while improving animal health and animal welfare. Nowadays, the following methods are employed:
\begin{itemize}
    \item RFID subcutaneous chips or rumen bolus, which can be read using a dedicated electronic reader;
    \item Ear tags, holding the animal identification number according to the country legislation format;
    \item Brand code, marked on the animal skin and used as a traditional identification system mainly in developing countries.
\end{itemize}
Authors from \cite{evans2005livestock} reported further details on RFID and electronic identification system for cattle in the US market, while the reader may refer to \cite{bowling2008identification} for an extensive review of cattle identification systems worldwide.

\subsection{Machine Learning Motivations and Insights}
The methods reported above suffer from major drawbacks. In particular, the use of RFID devices entails a significant cost for farmers of developing countries, because of the installation fees and the need of electronic readers during re-identification procedures. Moreover, they may, sometimes, have a sensible impact on animal welfare. On the other hand, ear tags are cheaper to buy but can be easily counterfeit or even removed through ear excision, beside being often lost by the animal itself. Consequently, there is room and a strong need to develop new methods with the following requirements:
\begin{itemize}
    \item Cheap for both installation and maintenance, including the supporting hardware;
    \item Able to be easily and rapidly used in real scenarios, as instance in the field or in a stable;       
    \item Hard to be counterfeit or removed.
\end{itemize}
Re-identification based on images holds all these properties, and can be exploited using the latest techniques and advances from Deep Learning and Computer Vision. Differently from traditional machine learning, deep learning techniques do not require any human hand-crafted features as they learn those representations directly from data, identifying features that may be more robust to pose or backgrounds variations as well as illumination changes. This is especially needed for cattle, since it is not easy to obtain images with a predefined pose of the animal, as it tends to move constantly while roaming or eating.

While for humans it is widely recognised that the face holds a great importance for visual identification purposes, in the animal kingdom a similar certainty still lacks, as almost no studies for this specific task have been conducted yet. Cattle present a high inter-breed variance in both body proportions and skin textures. On one hand, this makes fairly easy to distinguish cows of different breeds even for novices, on the other hand, due to the genetic selection made by humans in the past centuries and even more in the last decades, cows present a lower inter-breed variance compared to humans. However, despite this quite high degree of inbreeding, many cow breeds hold a unique texture pattern that is different from animal to animal, while also behaviour and social interaction contribute with marks, scratch and other defects that remain evident on the animal skin. 

In this work we used pictures of the head of cows taken from different angles of rotation and inclination, essentially for the following reasons:
\begin{itemize}
    \item The head of a cow shows a sufficient characteristic set of textures, shapes and patches. Even for textures-less breeds (such as the Bruna Alpina), the presence of horns and their length or the fur colour contributes to this variety. Furthermore, \cite{gleerup2015pain} showed how cattle face muscles are sufficiently developed to exhibit different facial expression, that may be used to distinguish one from another.
    \item Most of our images have been collected in farms with cows restrained during veterinary procedures, and only the head is easily accessible;
    \item Pictures of the full cow would introduce variance in both the animal pose and the background and could possibly require even more images of the same animal to perform the re-identification task.
\end{itemize}
Furthermore, an approach based on "facial" images could be compared with the current literature available on human re-identification.

\subsection{Main contributions and Novelties}
The contributions of this research are two-fold. Firstly, we provide a deep learning based framework for cattle re-identification. Secondly, we demonstrate how the use of multiple views of the same cow leads to superior performance w.r.t. standard approaches, the latters typically using only the front view image of the subject.

\subsection{Paper Structure}
The rest of this paper is structured as follows. Section \ref{sec:related} presents re-identification methods on both human and animal, focusing in particular on cattle. Section \ref{sec:proposed} introduces our proposed method, detailing the base block of our CNN architecture. Our cattle dataset is introduced and described in \ref{sec:dataset}. Section \ref{sec:experiments} shows extensive experiments and comparison with the baselines, while also discussing their performances. Section \ref{sec:ablation} further investigates our proposed architecture and its parts function. Finally, section \ref{sec:conclusions} summarises the contributions and results of the article.

\section{Related Works \label{sec:related}}
\subsection{Human Re-identification}
Human Re-identification has a long history of both research and practical uses. Among early methods, EigenFaces \cite{turk1991eigenfaces} has proved to perform well on cropped and aligned faces, such as the Olivetti dataset \cite{olivetti}, where it achieves a re-identification accuracy of 95\%. FisherFaces \cite{belhumeur1997fisher} employed a classifier based on the Linear Discriminant Analysis (LDA), exploiting features coming from a preprocessing phase involving a Principal Component Analysis (PCA) stage. In this way, it merges information from multiple views of the same subject in the final classifier.
However, both these methods are unable to deal with unaligned faces and suffer from illumination changes.

Since they are widely known for extracting more invariant features, Local Binary Pattern Histograms (LBPH) \cite{ojala1994performance}, Histogram of Oriented Gradients (HOG) and Scale Invariant Features Transform (SIFT) \cite{lowe1999object} descriptors have been widely used. As some of the more extend variations usually occur in illumination changes and scale, these descriptors have been designed to be robust for these applications. However, the human face has more than 15 muscles producing some of the most complex expression in nature, which can alter dramatically the final appearance of a person face.
In the latest years, DCNN trained on huge datasets, such as \cite{liu2015faceattributes},\cite{guo2016ms} and \cite{LFWTech} have provided features learned directly from examples with a growing interest from the computer vision and machine learning communities. Among them, \cite{schroff2015facenet} and \cite{liu2017sphereface} show state-of-the-art performances, as they can handle different facial expressions as well as face aging. Even with some differences, both architectures produce a low-dimensional feature vector, namely an embedding of the input face, efficient to be compared with others using Nearest Neighbours classifiers.

\subsection{Animal Re-identification}
Nowadays, little efforts have been made for the animal Re-identification task, with the noticeable exception of apes, since they share common traits with human beings. \cite{crouse2017lemurfaceid} achieved 98.7\% on facial images coming from 100 red-bellied lemurs (Eulemur rubriventer) of the Ranomafana National Park, Madagascar. The authors employed multi-local binary patterns histograms (MLBPH) and Linear Discrimant Analysis (LDA) on cropped and aligned faces. The authors show how human faces recognition algorithms can be adapted to primates, as their faces share the same underlying features.
Following their work, \cite{deb2018face} expanded the re-identification task to multiple apes species using DCNN. On one hand, the authors tried some traditional baseline such as EigenFaces and LBPH, on the other hand, they exploited two state-of-the art human faces identification networks, namely FaceNet and SphereFace.
Using the last, the authors showed how a CNN trained for solving a human re-identification task may also be adapted to the primates one, leveraging a fine-tuning strategy. However, they achieved slightly superior performance by means of a smaller CNN (in terms of number of layers and parameters), which is trained from scratch on apes' faces from three different species.
 In both above mentioned approaches, the underlying assumption lies on the presence of some similarity between the human and the ape faces. This evidence has been corroborated by two surprising results: firstly, the network trained on human faces perform enough well also on apes (showing comparable results with a network trained from scratch on apes \cite{deb2018face}) and secondly, it is able to extract and work with facial landmarks.

Other endangered species have also attracted interest in the latest years, from zebras \cite{lahiri2011zebra} to tigers \cite{hiby2009tiger}.\cite{norouzzadeh2018automatically} proposed a deep learning technique aiming to automatically identify different wildlife species, as well as counting the occurrences of each species in the image. Differently from us, such methods work on images depicting the entire animal's body, exploiting the characteristic stripe patterns of such animals.

Finally, pets represent an opportunity to build larger dataset, as they outnumber the above mentioned animals of a large margin, but public datasets still do not exist, and collecting this data requires huge resources and efforts. As an example, using pictures of two dogs breeds gathered from Flickr, \cite{moreira2017my} achieved remarkable performances. As dog faces differ from humans ones, the authors developed two Deep CNN trained from scratch on dogs images only, after a pre-processing phase consisting of a tight crop to suppress most of the background.

\subsection{Cattle Re-identification}
Due to their economic value, a literature regarding cattle-identification methods is slowly stemming in the latest years. However, to the best of our knowledge, this is the first attempt in doing it using the animal face with Deep Learning techniques.

\cite{andrew2017visual} employed images from Unmanned Aerial Vehicle (UAV) to identify cattle of a single breed using individual stripes and patches. Firstly, the authors gathered a dataset of 89 different cows depicted in 980 RGB images, the latters being captured by a camera placed over the walkway between holding pens and milking stations of a farm. Secondly, the authors presented a CNN trained from images, as well as a complete pipeline involving a Long-Short Term Memory (LSTM) layer to exploit temporal information. They achieved 86.07\% identification accuracy on a random train-test split and 99.3\% detection and localisation accuracy. 

Similarly, \cite{zin2018image} developed a system based on histograms and movements to record images of the backs of 45 cows from a camera placed on the Rotary Milking Parlour, for a period of 22 days. The authors trained a DCNN to perform individual identification, achieving an outstanding 98.97\% of accuracy. The collection system was able to correctly detect the back of the cow and crop it from the image. Using this approach, the system was able to record a huge amount of data with a great variation of light condition.

Since in the above described methods cows' pictures are taken from above, they are not suited and comparable with our scenario. Indeed, both approaches would require the constant presence of an expensive UAV for image acquisition or several difficulties for an operator. Therefore, their use could be extremely limited especially in developing countries. Furthermore, for textures-less animals, a picture of the back holds little details compared with one of the face, where traits such as the eyes and the muzzle vary greatly.

\section{Proposed Method \label{sec:proposed}}
In order to leverage the textures and details of both cattle profiles (i.e. the frontal and one of the two sides), we built an embedding DCNN starting from two images of the same cow.

At a high level, the network takes in input the two images and subsequently outputs a 128 dimensional embedding with unitary L2 norm. More in detail, each of the two images is independently processed by a separate convolutional branch, and their outputs are concatenated to form a single feature vector. It is worth noting that the two branches do not share any parameter, since different features may be required for the two animal profiles. 

\begin{figure}
\centering
\includegraphics[width=0.98\columnwidth]{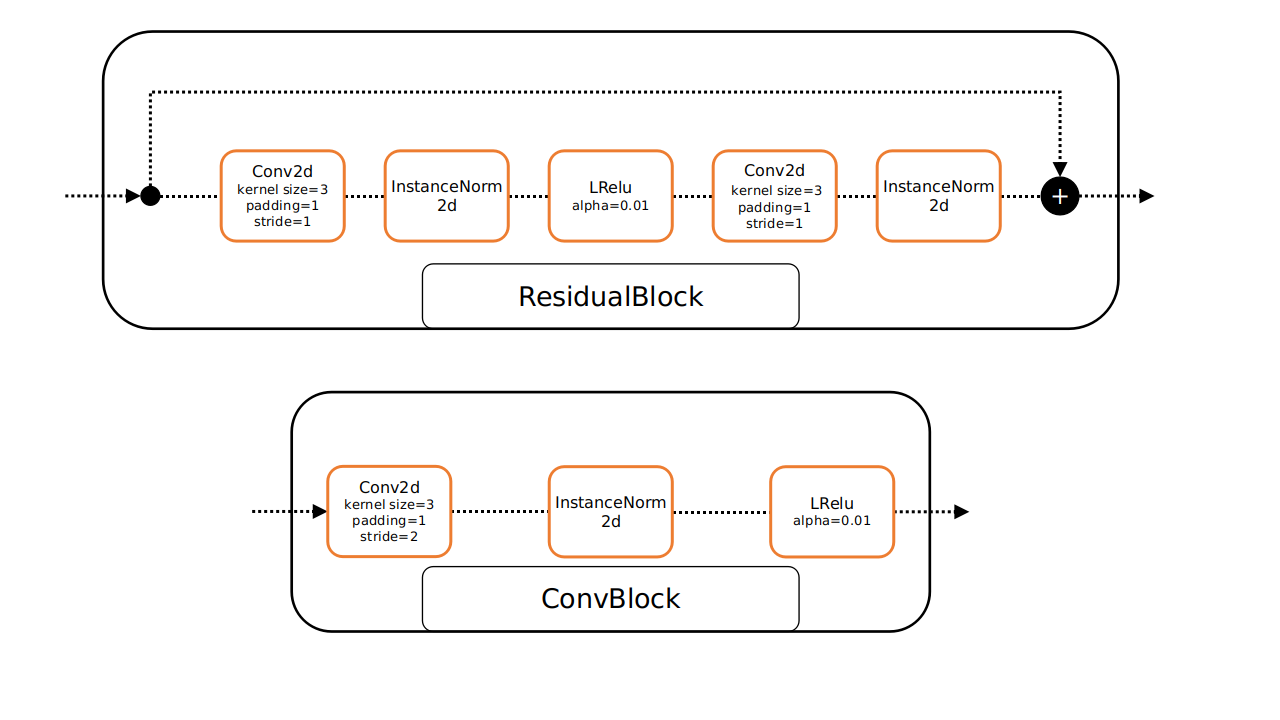}
\caption{Base blocks used in our architectures.}
\label{fig:blocks}
\end{figure}

Our multi-view network has been designed by means of two building blocks:
\begin{itemize}
    \item \textbf{ConvBlock}: A single Convolutional layer followed by InstanceNorm \cite{DBLP:journals/corr/UlyanovVL16} and LReLU activation, reducing the feature maps' spatial resolution by a half;
    \item \textbf{ResBlock}: A residual unit \cite{he2016deep} with LReLU activation, preserving the spatial resolution. 
\end{itemize}
A scheme of these blocks is shown in Figure \ref{fig:blocks}.

\begin{figure*}
\centering
\includegraphics[width=1.75\columnwidth]{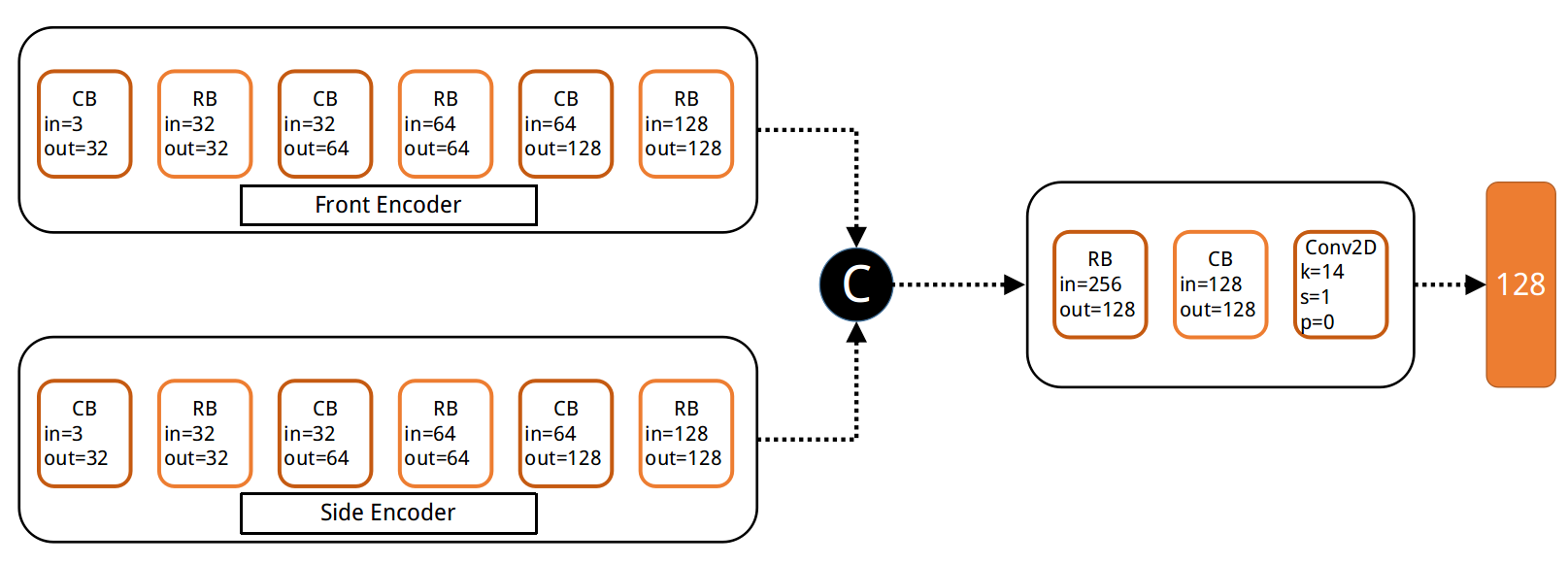}
\caption{Our multi-views architecture. Note that CB stands for ConvBlock, while RB for ResidualBlock, as described in Figure \ref{fig:blocks}}
\label{fig:arch}
\end{figure*}

 Instead of using the more popular batch normalisation, we address the internal covariate shift problem by means of instance normalisation. Indeed, with the second, we observed improvements in terms of stability during the training phase.
The network ends with a single 2D Convolution, with kernel size equals the feature map size and number of filters equals to the desired embedding size (i.e. 128). In this way, each input map is reduced to a single scalar value, leading to a fully convolutional architecture. The overall architecture is presented in Figure \ref{fig:arch}.

We employed the Histogram Loss from \cite{ustinova2016learning} as the only loss function of our architecture. 
After a batch of anchors, positives and negatives is embedded into a high-dimensional space by a deep network, the loss computes the histograms of similarities of positive and negative pairs. The integral of the product between the negative distribution and the cumulative density function for the positive
distribution is evaluated, corresponding to a probability that a randomly sampled positive pair has smaller similarity than a randomly sampled negative pair. We performed extensive comparison using the triplet loss function described in \cite{schroff2015facenet}, but found the latter more unstable during the train phase.

\section{Dataset \label{sec:dataset}}
A potential drawback in the use of Deep Learning methods is that a huge amount of pictures depicting cows’ heads is required to achieve reasonable performances on unseen examples. Furthermore, the training set should include a great variety of poses, illumination changes and background for each subject, with the acquisition process spanning potentially in multiple days. However, to the best of our knowledge, such dataset still does not exist for cows' faces. Thus, we collected pictures and video of cattle from four Italian farms distributed in three regions. We collected videos and extracted images from those for the training process, while employing only pictures acquisitions for the test phases. We leveraged the Vatic tool \cite{vatic} to annotate the cows' faces with a bounding box for each frame. Finally, we discarded some of the extracted frames aiming to ensure a high inter-frames variance. 

Such activity should be considered as mandatory, since the animal usually moves slowly during video acquisition, introducing a lot of redundancy if all the frames are kept. Moreover, a traditional setting for the re-identification task consists of few different pictures per single identity.

Eventually, we obtained the following splits:
\begin{itemize}
    \item Train Set; consisting of 12952 pictures from 387 different subjects;
    \item Database Set; consisting of 4289 pictures from 52 different subjects, recorded during two different days;
    \item Test Set; consisting of 561 pictures from 52 different subjects. These cows are the same included in the Database Set;
\end{itemize}
Some random samples from the last two set are shown in Table \ref{tab:samples}. It is worth noting that, given an image, one cannot make any assumptions regarding the cow's face location and orientation. Moreover, because of the oblong shape of cow faces, any alignment would lead to part of the cow face being cropped. Finally, only few landmarks detector work on animal faces \cite{rashid2017interspecies}, but they need to be fine-tuned on the animal domain, thus requiring expensive landmarks annotations.

\newcommand{\cowimg}[1]{\includegraphics[width=0.2\columnwidth,height=0.2\columnwidth]{imgs/cows/cow_#1.jpg}}

\begin{table}[]
\begin{tabular}{llll}
 \cowimg{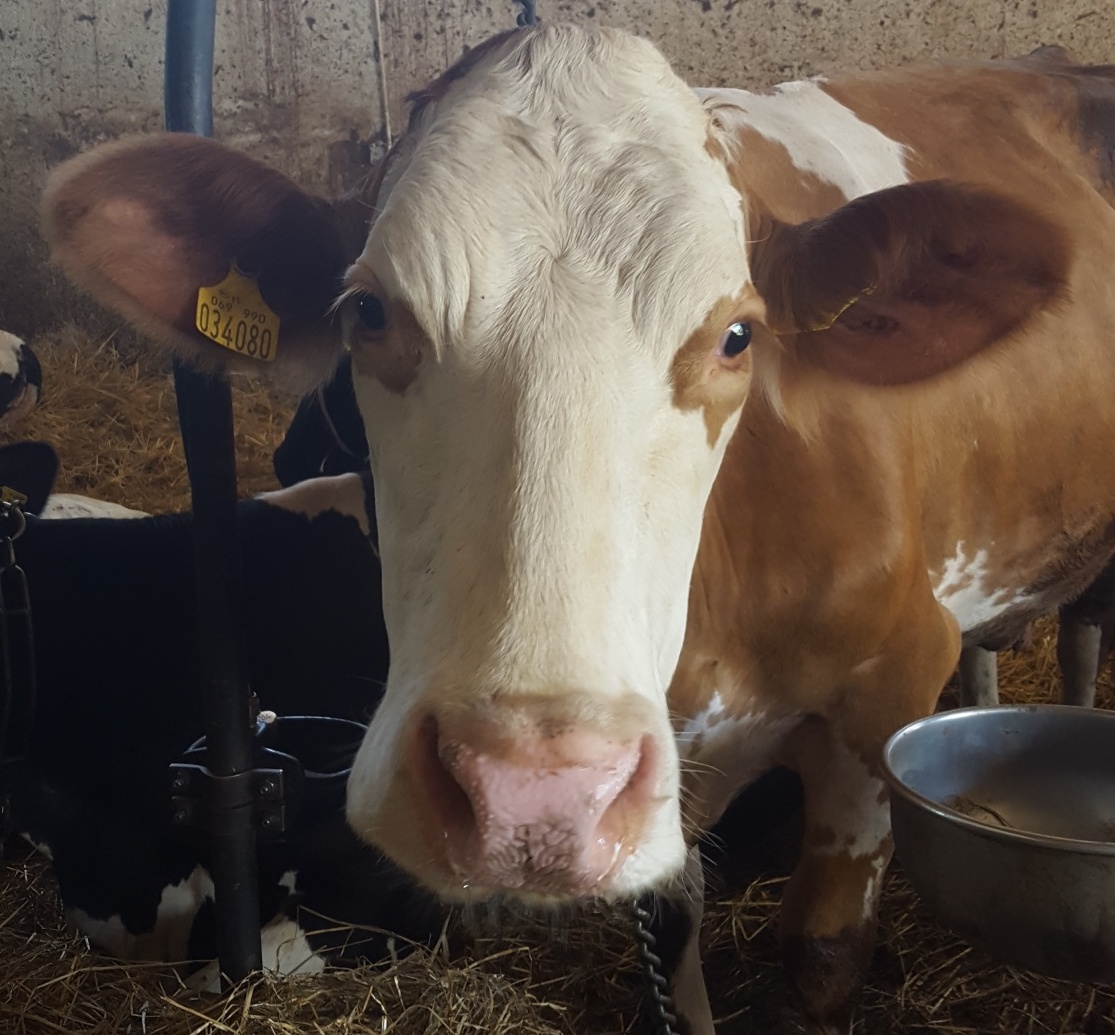} &  \cowimg{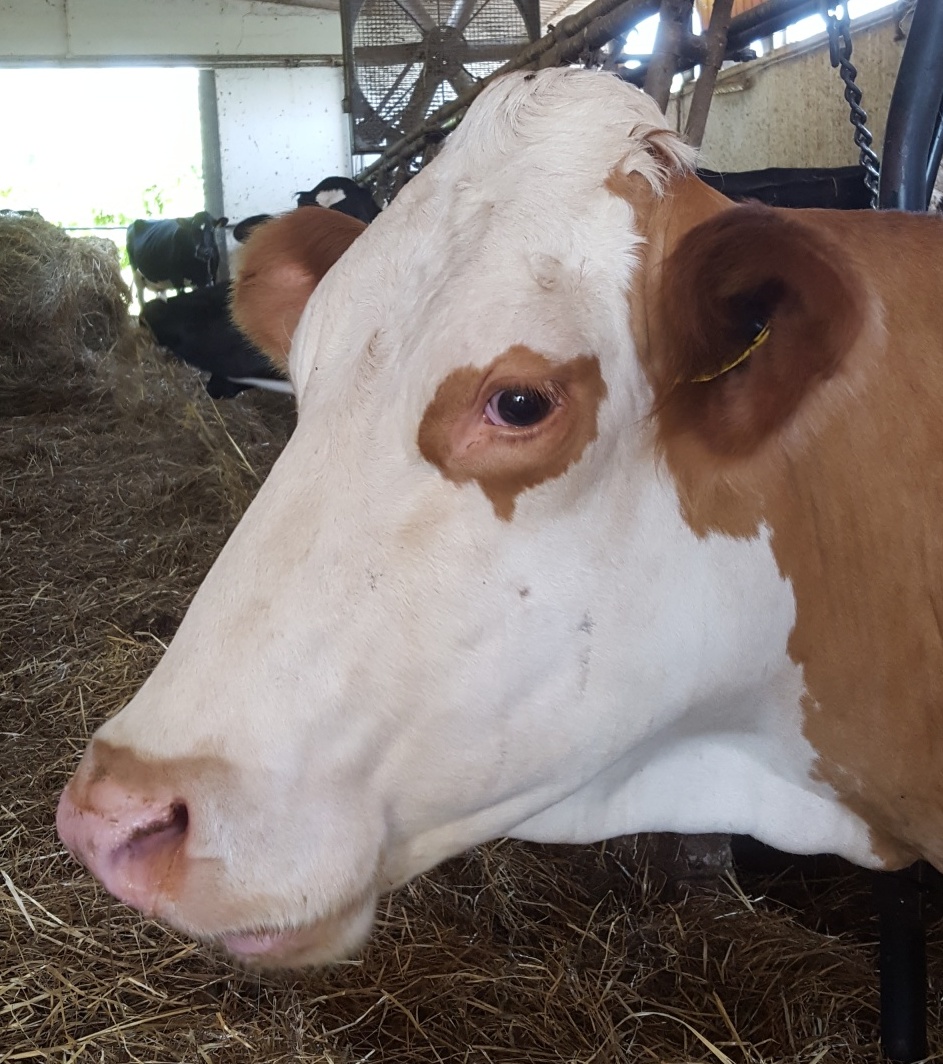} &   \cowimg{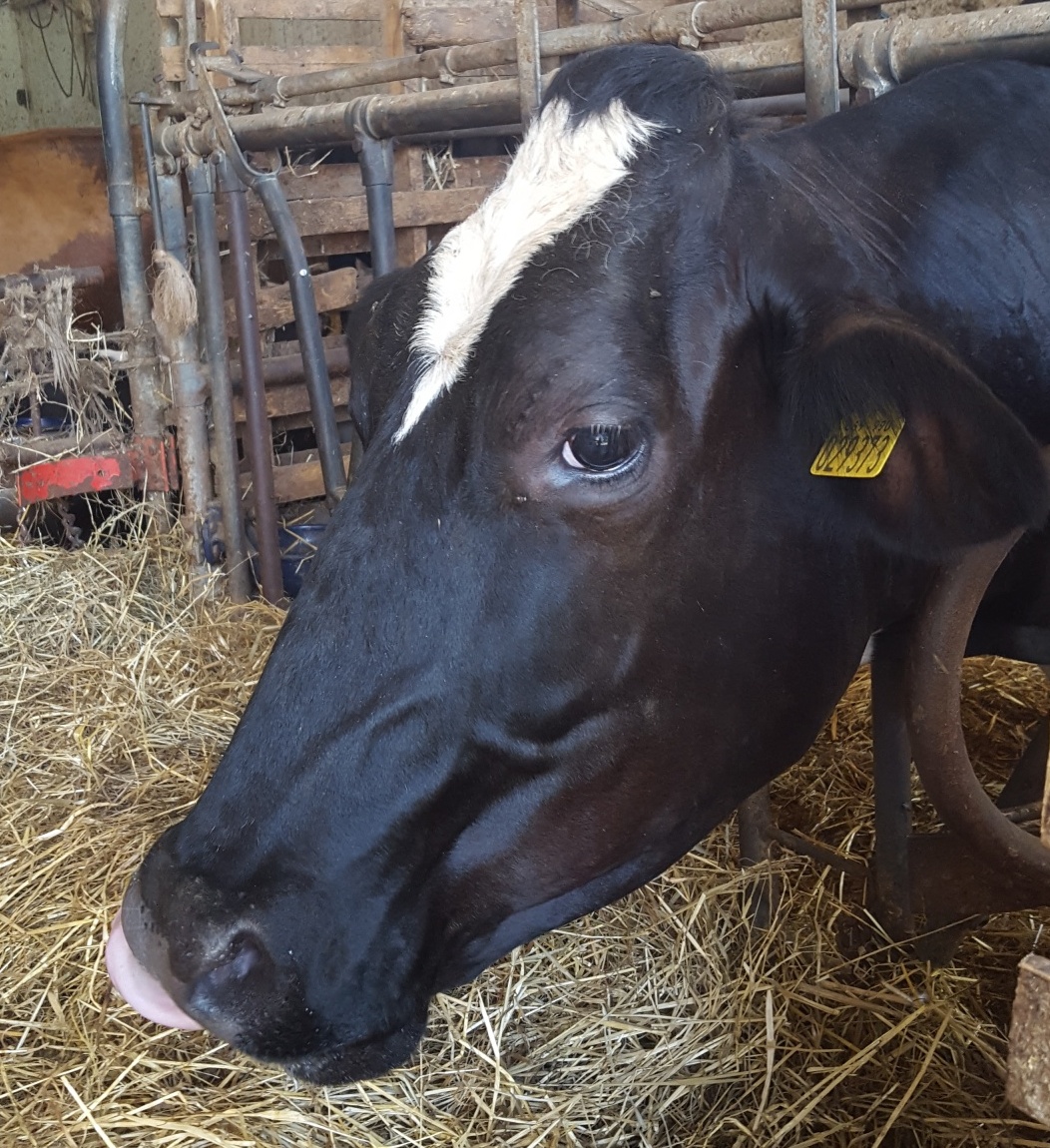} &  \cowimg{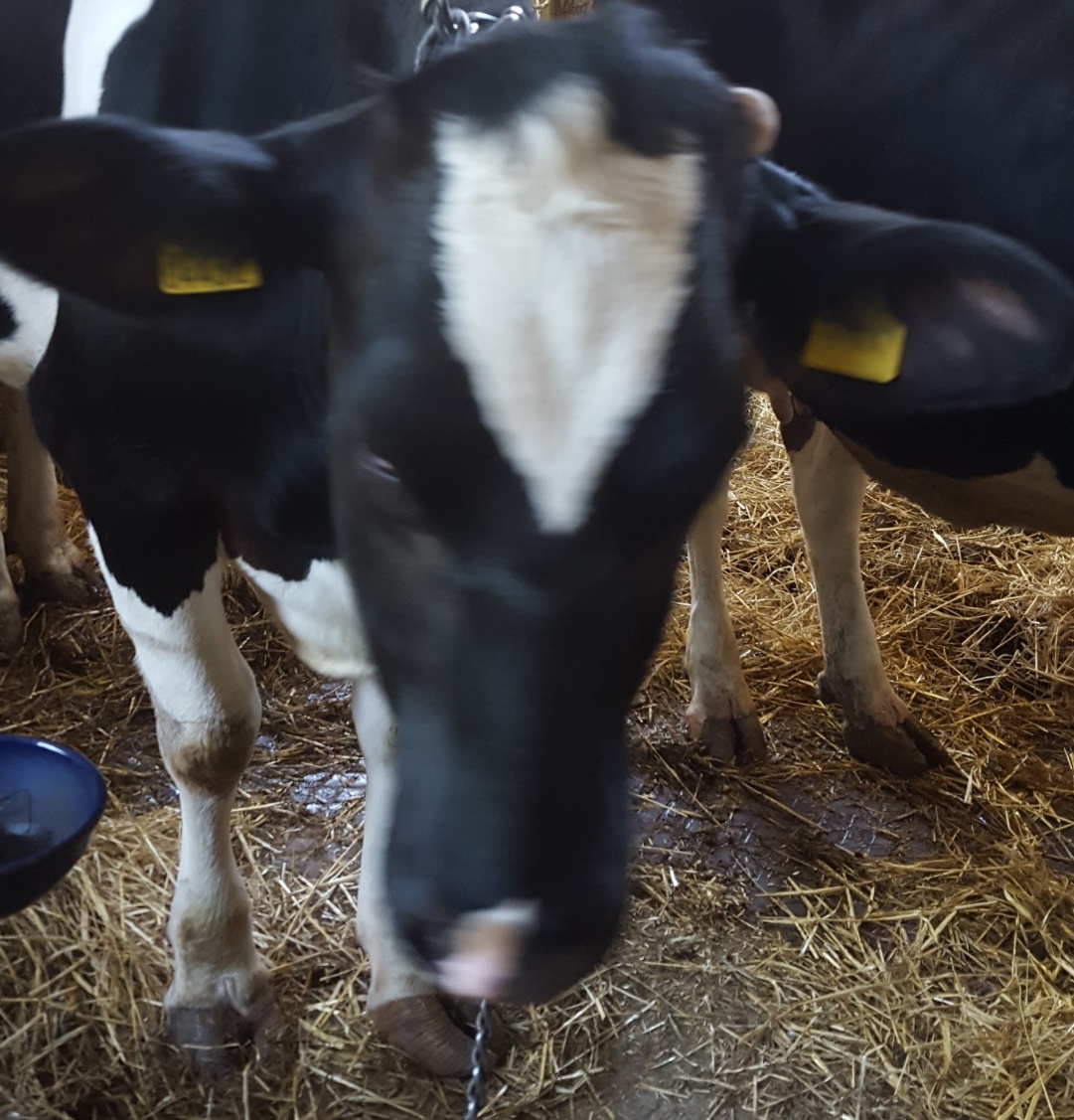} \\
 \cowimg{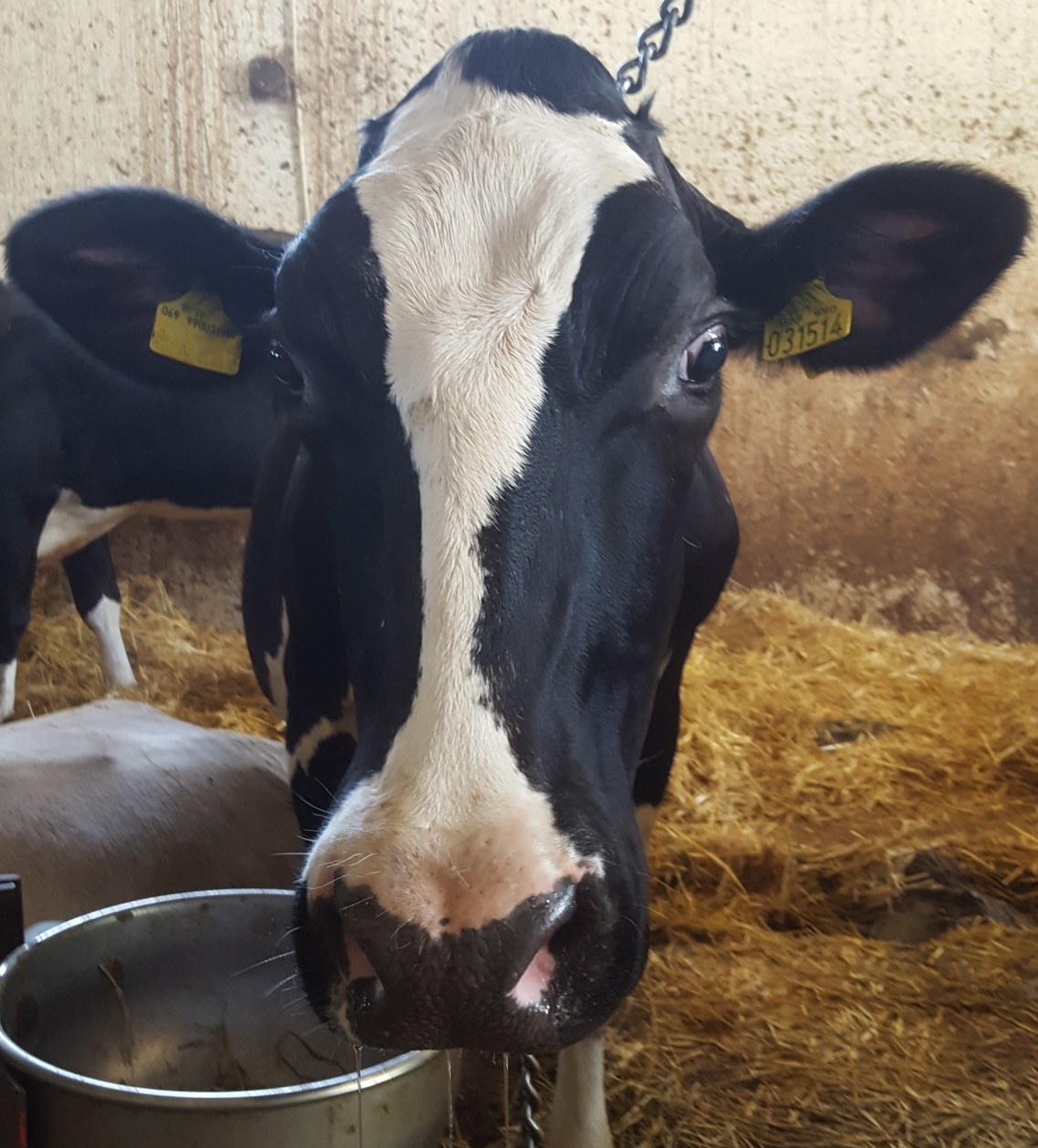} &  \cowimg{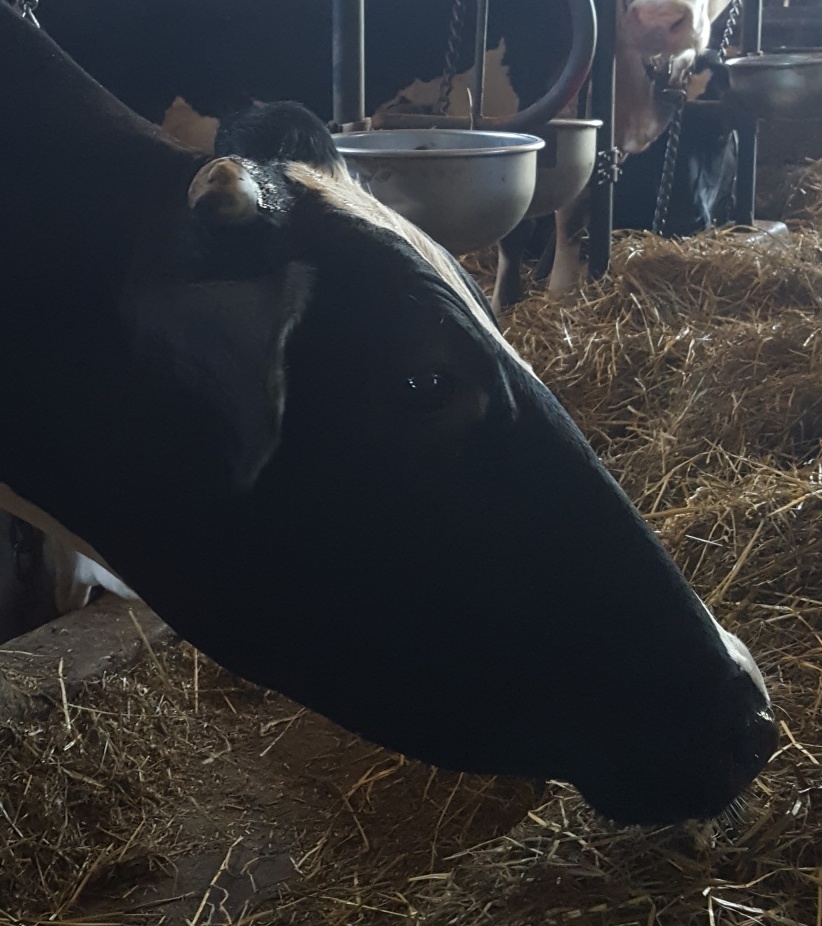} &  \cowimg{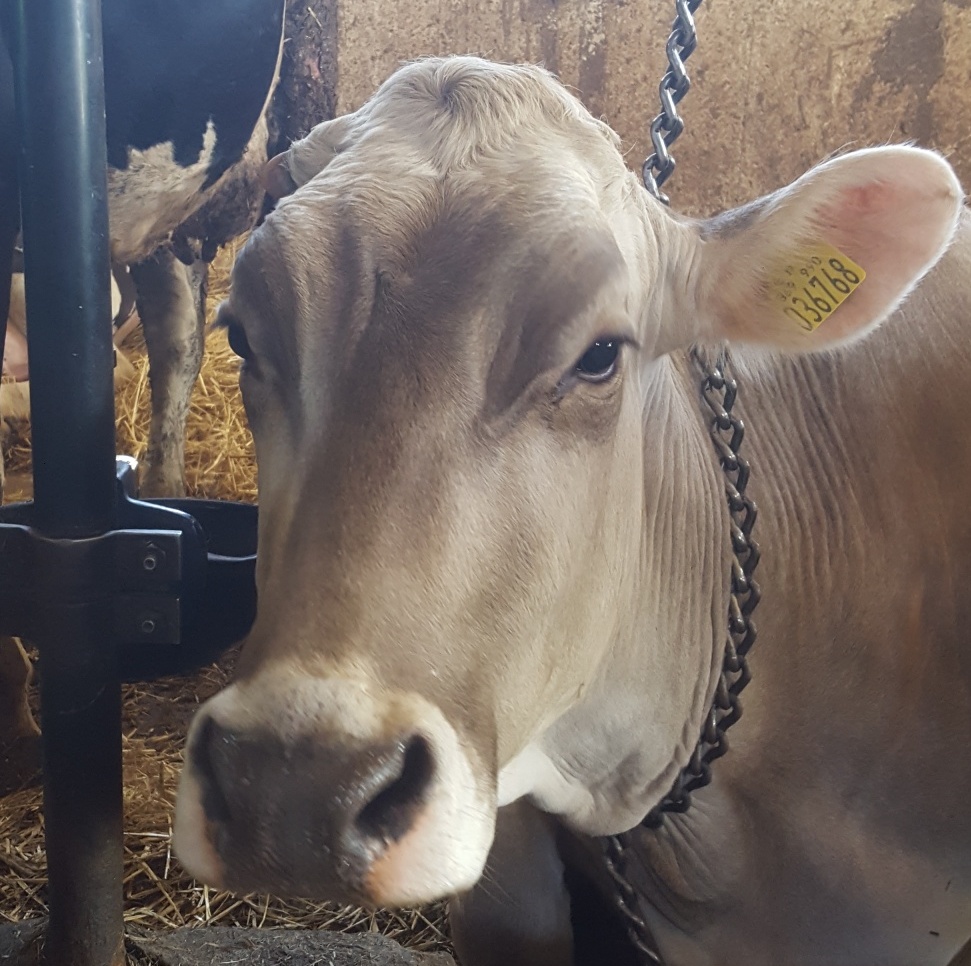} & \cowimg{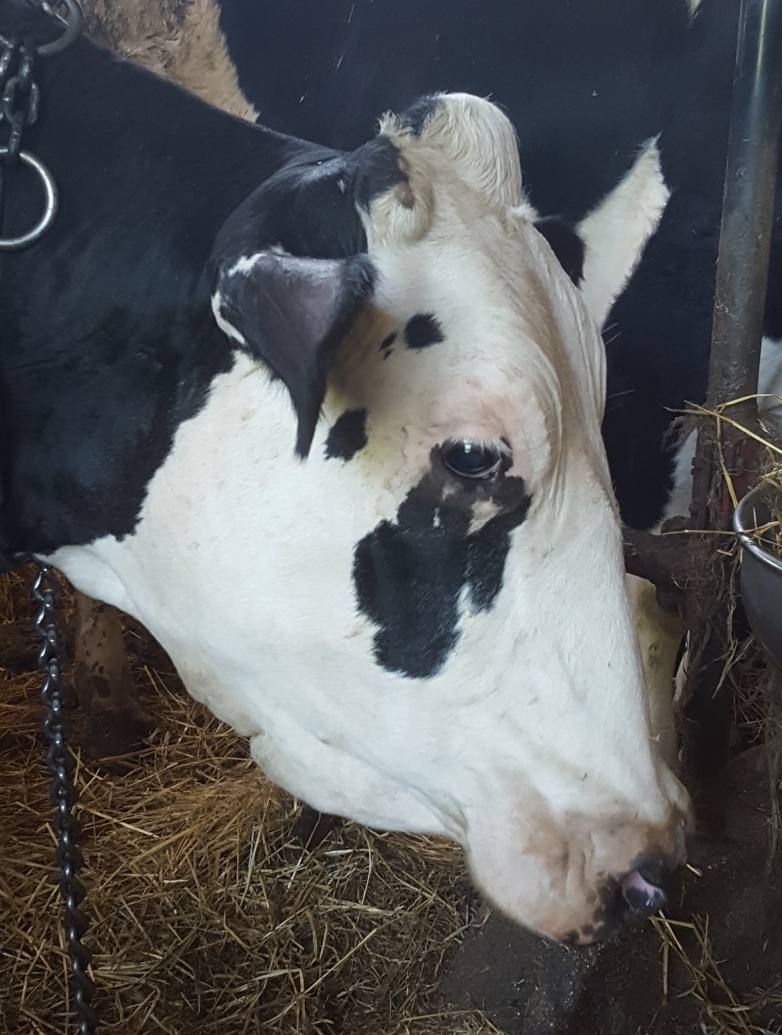} \\
  \cowimg{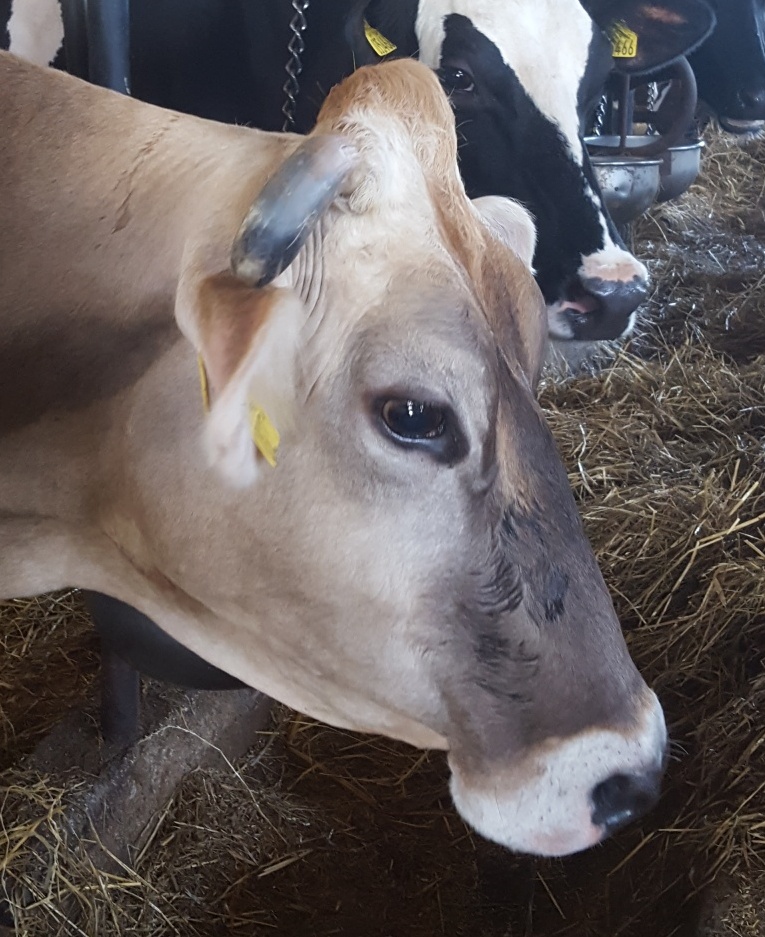} &  \cowimg{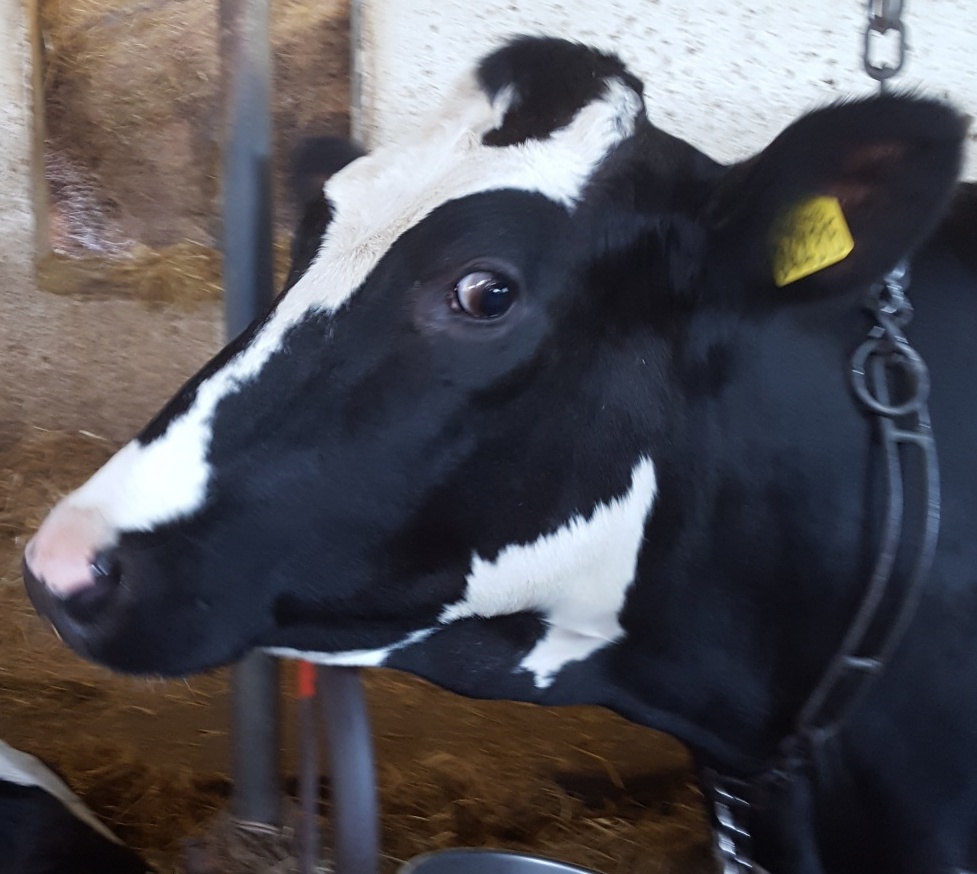} &  \cowimg{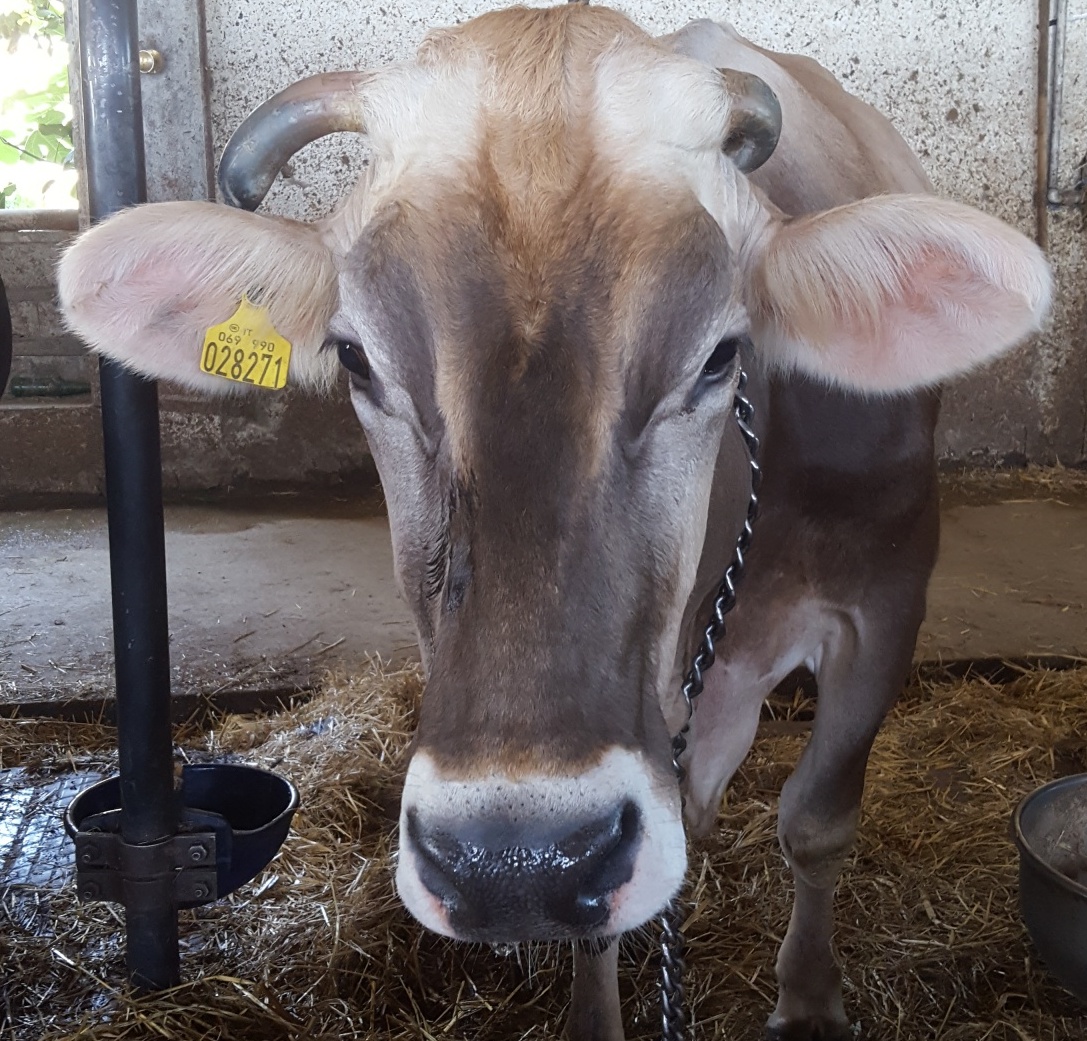} & \cowimg{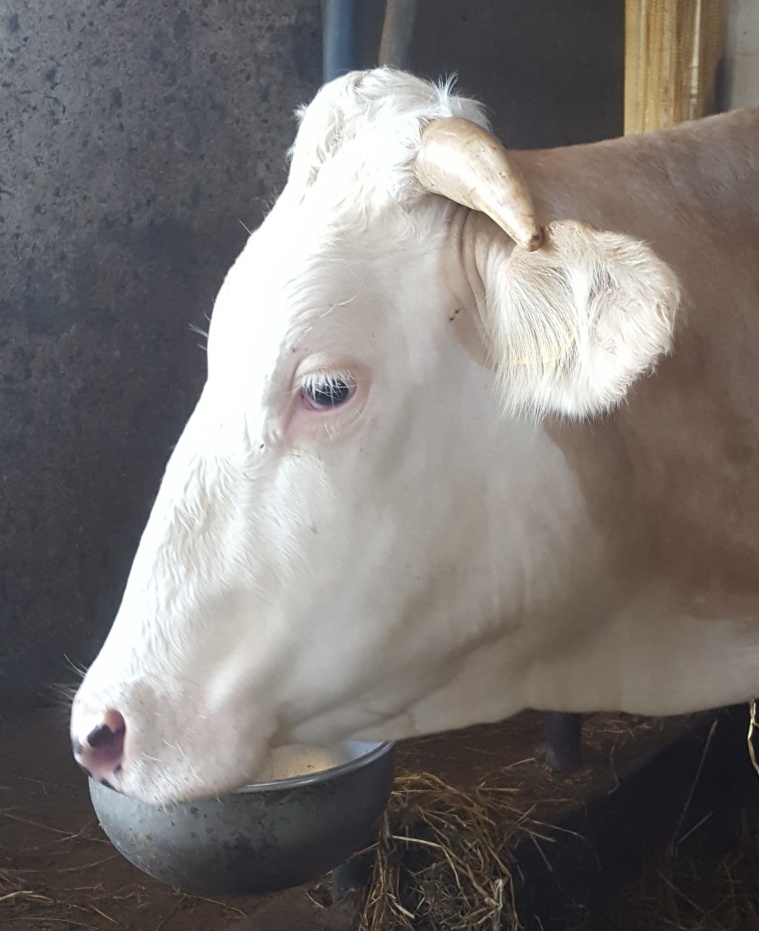} \\
   \cowimg{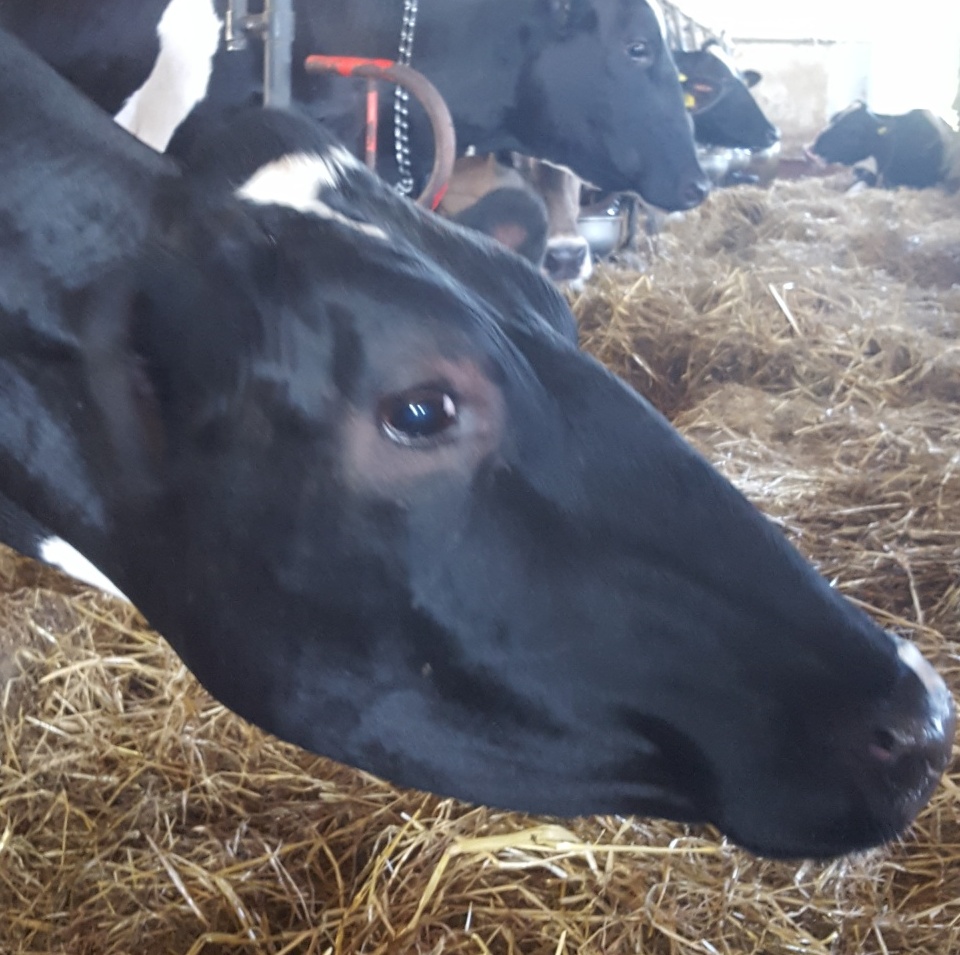} &  \cowimg{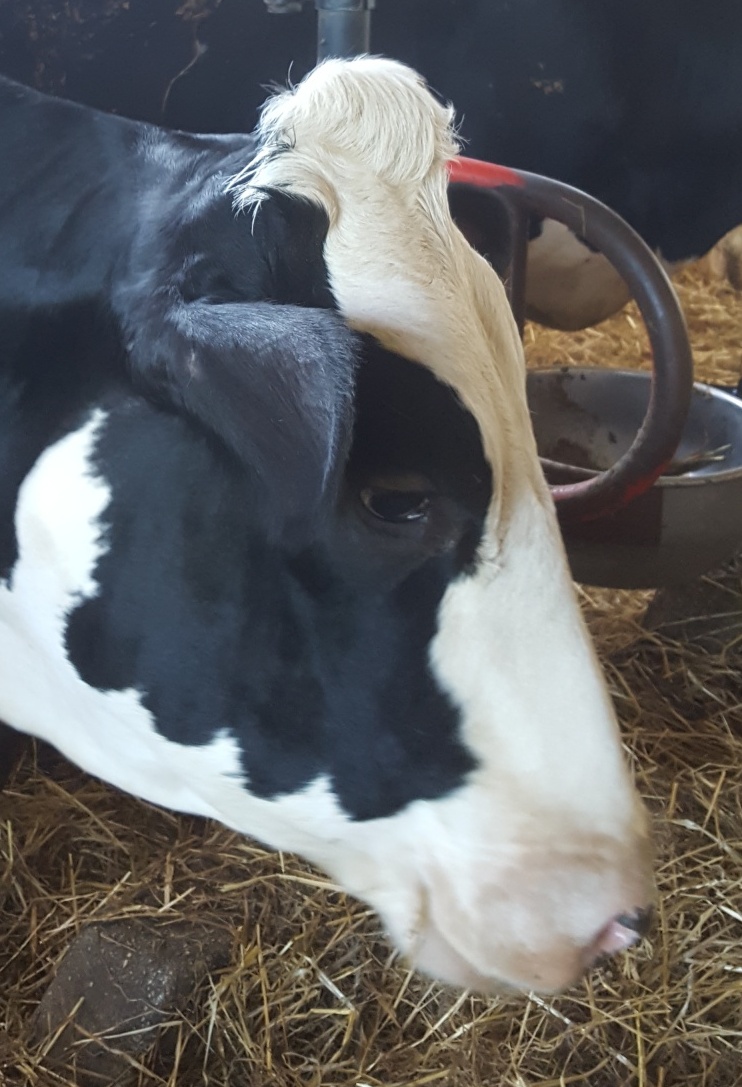} &  \cowimg{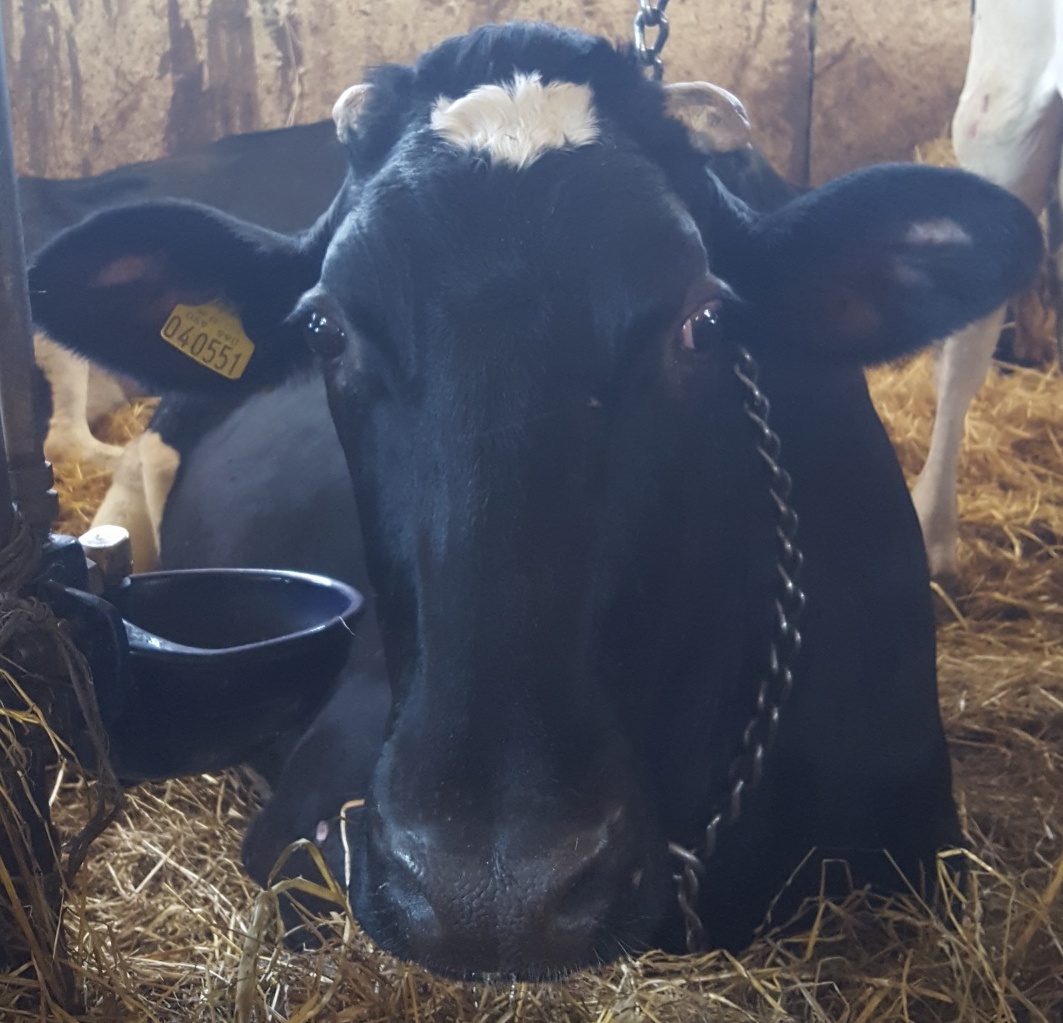} & \cowimg{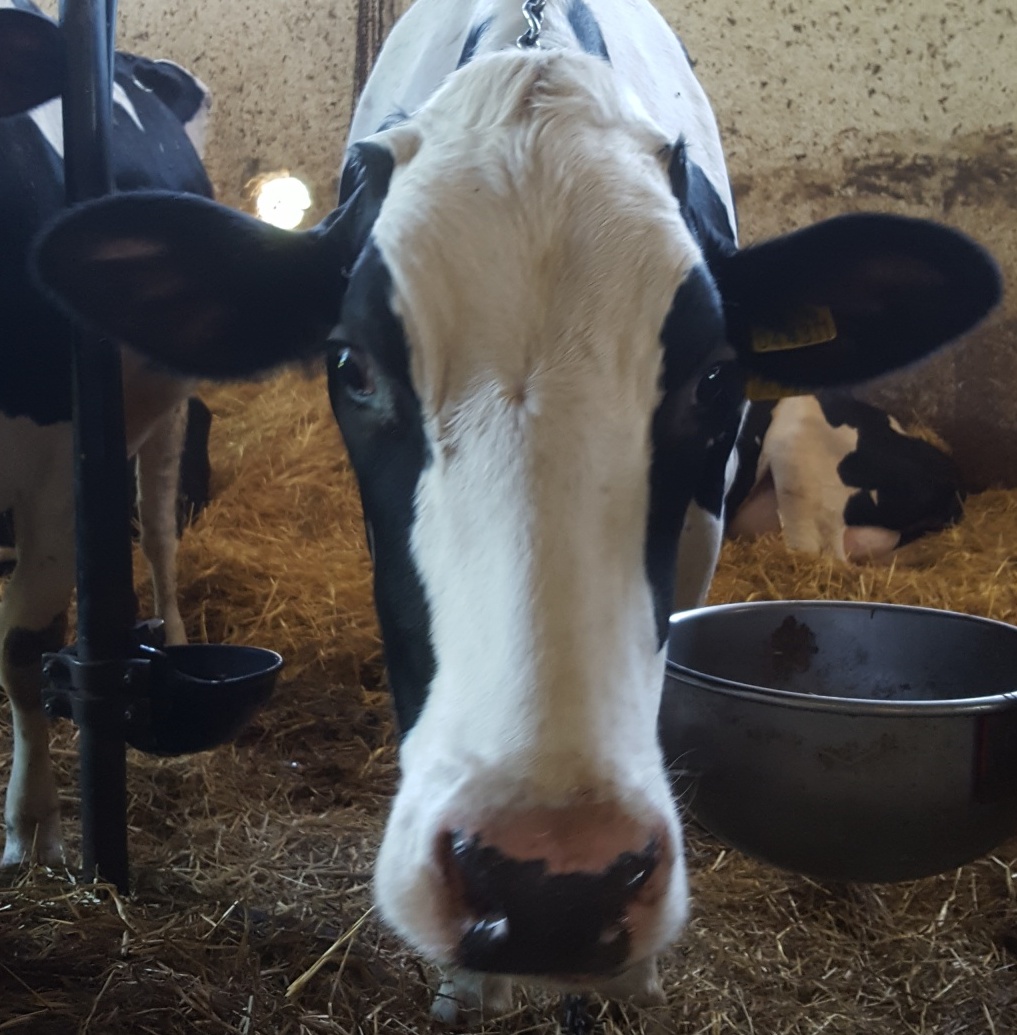} \\
\end{tabular}
    \caption{Some randomly drawn samples from our dataset.}
    \label{tab:samples}
\end{table}

\section{Experiments \label{sec:experiments}}

\subsection{Metrics}
Following \cite{deb2018face} and \cite{liu2017sphereface}, we test our solution in two different settings:
\begin{itemize}
    \item \textbf{Open-Set}: Identities of the test images are included in the train set.
    \item \textbf{Closed-Set}: Identities of the test images are separated from the train set ones.
\end{itemize}
Regarding the last, we consider it as more challenging and general, being also able to provide a good estimation of the generalisation capacity of the proposed model. For both settings, we conducted the same experiment, namely the \textbf{Identification}. Given images and correspondent ground-truth identities from a test set, the matching strategy returns the k nearest neighbours from the database set.
It is worth noting that the above mentioned "matching strategy" can be implemented by every classifier.

\subsection{Baselines \label{subsec:baselines}}
We include both deep and non-deep baselines to further present and motivate the main challenges of this novel task. For non-deep baselines, we include methods traditionally employed by the computer vision community for the human re-identification task. The reason behind this choice is not only to emphasise the differences between the human related task and the cattle one, but also because the great majority of them are provided as open-source verified software with the open-cv \cite{opencv_library} package. 

EigenFaces \cite{turk1991eigenfaces} method consists of a Principal Component Analysis (PCA) applied to images of human faces. The first $k$ eigenvectors, sorted by their respective eigenvalues magnitudes, can be seen as prototypes used to build the data. Test images can then be projected to extract $k$ coefficients, each one describing how much a prototype contributes to the image.

FisherFaces \cite{belhumeur1997fisher} extends Eigenfaces by means of the Fisher Discriminant Analysis (FDA), aiming to force multiple images of the same identities to lie in a nearby region of the subspace. While PCA preserves maximum variance during the projection, FDA attempts to preserves the discrimination capability at the end of such transformation. Indeed, FDA may be consider a supervised embedding algorithm, since it finds a projection that maximises the scatter between classes and minimises scatter within classes. Thanks to this, the Fisherface method shows superior performances w.r.t. Eigenfaces under the presence of variation in lighting and expression. 

LBPH\cite{ojala1994performance} has been widely used as a texture description. It builds a circular neighbour with a certain radius for each pixel, and extracts features based on the relationships between each pixel and its neighbour. The latters are carried out by a histogram, which is then used to describe the image, and can act as a feature descriptor for a further PCA.

HOG, similarly to LBPH, builds a histogram using neighbours but, instead of pixels values, the spatial derivatives are employed. The histogram takes into account both the magnitude and the orientation of the gradients, with the latter being quantized to contribute to achieving invariance to orientation.

The authors of SphereFace\cite{liu2017sphereface} present a DCNN trained on a large human faces dataset, achieving state-of-the-art performances in an open-set setting. The images are firstly aligned and cropped, and a 512-dimensional feature vector is extracted from the second-last layer of the network. A classification loss, named Angular Softmax, is proposed: on one hand, it requires examples from the same identity to lie nearby on the output landscape. On the other hand, it forces examples from different identities to be spaced by a considerable margin, the latter being in the form of an angle on a hypersphere. As versions of the network pretrained on human faces are available, results with and without a training phase on cattle are reported in subsection \ref{subsec:res}. 

It is worth noting that all methods listed above share the same output representation; in particular a feature vector (embedding) is produced from a given input image and, as such, the "matching strategy" can be the same for all of them. 

\subsection{Implementation Details \label{subsec:details}}
As far as it regards the train methodology, it is worth noting that:
\begin{itemize}
    \item For the \textbf{closed set} scenario we train on both the Train and the Database Set, while for the \textbf{open set} one we train only on the first;
    \item We pre-processed the images by scaling them to a fixed size (i.e. 224). 
    \item We performed data augmentation by randomly rotating, cropping and projecting images, while also changing the hue and saturation of the images. We didn't perform horizontal flip, as it causes a noticeable drop in performances;
    \item We employed the Histogram Loss using a batch size of 64 triplets and 200 bins for the histograms;
    \item We mined both hard positives and negatives during the training phase. The firsts are positives with an embedding extremely far away from the average embedding of the identity, while the lasts are the closer negatives to the identity in the embedding space;
\end{itemize}

\begin{table*}[h!]
\centering
    \begin{tabular}{c | cc cc cc cc cc cc cc}
    \hline
    & \multicolumn{2}{c}{EigenFaces} & \multicolumn{2}{c}{FisherFaces} & \multicolumn{2}{c}{LBPH}  & \multicolumn{2}{c}{HOG} & \multicolumn{2}{c}{SphereFace} & \multicolumn{2}{c}{SphereFace(cattle)} & \multicolumn{2}{c}{Ours}\\
    & Open & Close & Open & Close & Open & Close & Open & Close & Open & Close & Open & Close & Open & Close\\\hline
    Top1 & 0.227 & 0.229 & 0.242 & 0.237 & 0.263 & 0.265 & 0.39 & 0.4 & 0.139 & - & 0.367 & 0.556 & 0.558 & \textbf{0.817}\\
     Top3 & 0.352 & 0.354 & 0.345 & 0.345  & 0.406 & 0.415 &  0.522 & 0.532 & 0.226 & - & 0.46 & 0.636 & 0.742 & \textbf{0.891}\\\hline
     
    \end{tabular}
    \caption{Results for the Identification task.}
    \label{tab:results}
\end{table*}

\begin{figure*}[h!]
\centering
\includegraphics[width=1.8\columnwidth]{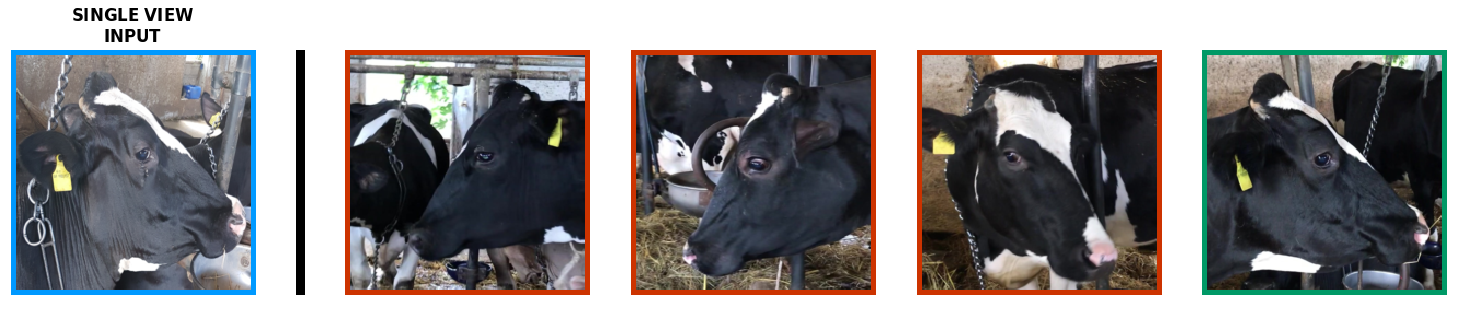} \\
\includegraphics[width=1.8\columnwidth]{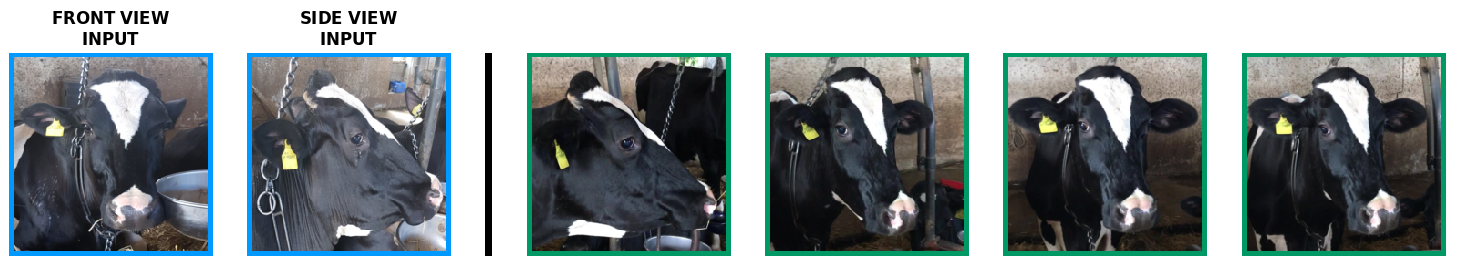}\\
\vspace{0.10\columnwidth}
\includegraphics[width=1.8\columnwidth]{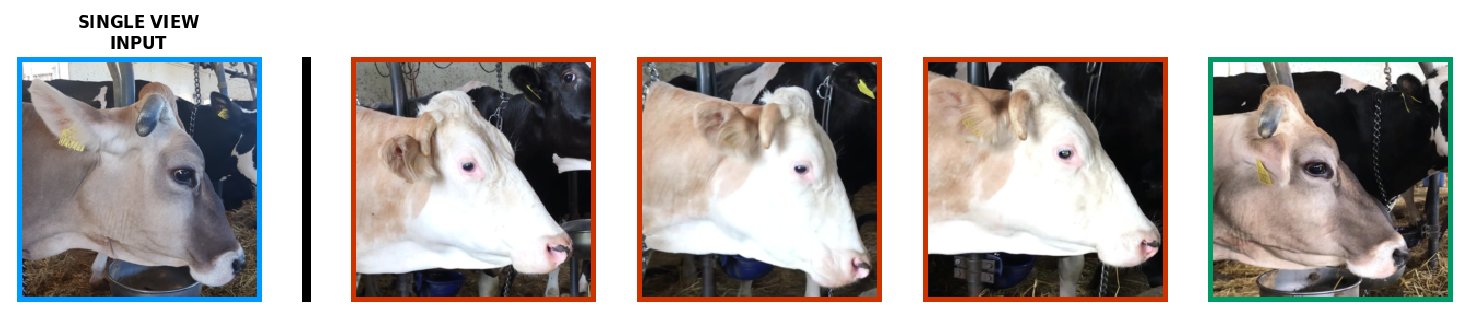} \\
\includegraphics[width=1.8\columnwidth]{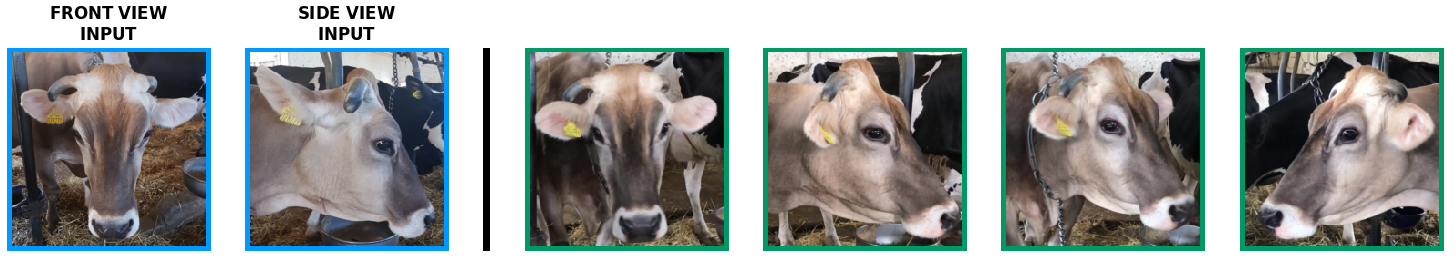} \\
\caption{Illustration of the K-NN retrievals, computed on two individuals using a single view or a double view architecture. Input images have been labelled with blue contour, while green and red have been used respectively for correct and misclassified examples. Best view in colours. }
\label{fig:single_multi}
\end{figure*}

\subsection{Results \label{subsec:res}}
As shown in Table \ref{tab:results}, we compare our results with the baselines discussed in \ref{subsec:baselines}. For each method, the hyper-parameters tuning activity has been conducted using a grid-search strategy. For the identification task, results are reported in terms of Top1 and Top3 match, for both open and close settings. For every methods, KNN (k-nearest neighbours) has been employed as final classifier, as it requires no hyper-parameters grid-search nor supervised training, while also scaling linearly with the number of identities.

The input of each method is a single image, whereas our proposed method can be provided with two different profiles images. In order to enable a fair comparison and to motivate our design choices, results with only one profile image are reported in subsection \ref{subsec:multi}.

Looking at results showed in Table \ref{tab:results}, the following conclusion may be drawn: 
\begin{itemize}
    \item Methods based on local and stationary property (i.e. LBPH, HOG and CNNs) achieve better performance than Eigenfaces and Fisherfaces, which do not implicitly exploit nearby pixel correlations or local pattern's presence. 
    \item Deep models trained on cattle performs better than shallow ones, due to their robustness to different poses and other major source of variations (i.e background or illumination changes).
    \item SpereFace, when trained on aligned human faces, does not generalise to cattle. Such result show how the cattle re-identification task has nothing to do with the same task in the human domain, differently from what happen with apes. 
    \item Our solution outperforms by a consistent margin all the other competitors, including a state-of-the-art human re-identification network as SphereFace (even if trained from scratch on cattle). Such improvement is achieved by leveraging two different cow's profile, which leads to a higher discriminative capability for similar subjects.
\end{itemize}

\section{Ablation Study \label{sec:ablation}}

\subsection{Multi-view vs Single-view \label{subsec:multi}}
Results for the Identification task reported in Table \ref{tab:results_multi_single} highlights the superiority of the multi-view approach over the single-view one. \ref{fig:single_multi} shows some cherry-picked example where the single-view model fails, while the multi-view approach effectively fuses features from both views to produce a more representative and robust embedding vector. 
This justifies the request of both views for a single prediction. Indeed, if two cows may have the same visual appearance under some poses or particular light conditions, on the other hand this possibility has a lower probability under the presence of both profiles. 

Moreover, cattle usually do not share a symmetry between left and right profile, as they often present very different spots and patterns. Also for such reason, the use of two profiles instead of one should be considered useful to find a meaningful discrepancy between two cows.

\begin{table}
\centering
    \begin{tabular}{c | cc cc }
    \hline
    & \multicolumn{2}{c}{Single-View} & \multicolumn{2}{c}{Multi-View}\\
    & Open & Close & Open & Close\\\hline
    Top1 & 0.443 & 0.688 & 0.558 & \textbf{0.817} \\
     Top3 & 0.575 & 0.748 & 0.742 & \textbf{0.891}\\\hline
    \end{tabular}
    \caption{Comparison between single and multi views methods.}
    \label{tab:results_multi_single}
\end{table}

\subsection{Extended Database \label{sub:ext}}

In Table \ref{tab:results_ext} we report results, in term of accuracy, showing how the use of the only database set during the test phase leads to better performances with respect to the union of the train and database sets. However, for the close set scenario, the drop of performance between the two settings is much lower w.r.t the open one, highlighting how the network improves its behaviour if the animal's images are available during the train phase. In this way, such knowledge may be used during the test phase to reject other subjects with similar patterns or characteristics. To further highlight the differences in terms of difficulty between the two settings a random predictor has been included.

\begin{table}
\centering
    \begin{tabular}{c | cc cc }
    \hline
    & \multicolumn{2}{c}{Database Set} & \multicolumn{2}{c}{Extended Set}\\
    & Open & Close & Open & Close\\\hline
         Random3 & 0.057 & 0.057 & 0.006 & 0.006\\
    Top1 & 0.558 & \textbf{0.817} & 0.396 & 0.732 \\
     Top3 & 0.742 & \textbf{0.891} & 0.583 & 0.820\\\hline

    \end{tabular}
    \caption{Comparison with different set during the test phase. Random3 stands for a random predictor scoring 1 if the correct identities lies among the first 3 prediction.}
    \label{tab:results_ext}
\end{table}

\section{Conclusions \label{sec:conclusions}}

In this work we propose a Deep Learning base method for cattle re-identification in unconstrained environment from single and multi-views. We present extend baseline comparisons both with non-deep and deep methods. We show that human and cattle re-identification are slightly similar tasks, but present important and significant differences. Finally, we highlight how a multi-views method (i.e a method combining information from multiple profiles) clearly outperforms both baselines and single-view methods.

\section*{Acknowledgment}
The authors thank Allevamento Martin, Allevamento Costantini, Cooperativa Venditti and Allevamento Hombre for providing access to their farms during the dataset acquisition. Moreover, the authors would like to acknowledge Francesco di Tondo, Lisa Leonzi Giuseppe Carolla and Carmen Sabia for their precious help with data acquisition.

{            
\bibliographystyle{IEEEtran}

}  

\appendix[Alignment Importance for human faces]
As already stated in \ref{sec:dataset}, alignment for animal faces poses major challenges in terms of annotation costs. Furthermore, since cattle posses oblong-shaped heads, an alignment using landmarks may lead to crops with deformed cow's proportions. To investigate the real impact of landmarks alignment, in terms of re-identification accuracy over human faces, we test the SphereFace architecture trained on CASIA-Webfaces on a split from the CELBA dataset, both with and without alignment. It is worth noting that even when the faces are not aligned, they are still cropped tightly around the subject face. Results from Table \ref{tab:results_align} show a noticeable drop of performance using non-aligned faces. Even if CNN should be robust to translations, other transformations such as rotations and prospective seem to affect performances.

\begin{table}[h!]
\centering
    \begin{tabular}{c | c c }
    \hline
    & SphereFace (Aligned) & SphereFace (Not Aligned)\\\hline
     Top1 & 0.976 & 0.390  \\
     Top3 & 0.978 & 0.565 \\\hline
    \end{tabular}
    \caption{Results raised comparing SphereFace both on aligned or not faces. }
    \label{tab:results_align}
\end{table}

\end{document}